# Improving Speech Related Facial Action Unit Recognition by Audiovisual Information Fusion

Zibo Meng, *Student Member, IEEE,* Shizhong Han, *Student Member, IEEE,* Ping Liu, *Member, IEEE,* and Yan Tong, *Member, IEEE*

**Abstract**—It is challenging to recognize facial action unit (AU) from spontaneous facial displays, especially when they are accompanied by speech. The major reason is that the information is extracted from a single source, i.e., the visual channel, in the current practice. However, facial activity is highly correlated with voice in natural human communications.

Instead of solely improving visual observations, this paper presents a novel audiovisual fusion framework, which makes the best use of visual and acoustic cues in recognizing speech-related facial AUs. In particular, a dynamic Bayesian network (DBN) is employed to explicitly model the semantic and dynamic physiological relationships between AUs and phonemes as well as measurement uncertainty. A pilot audiovisual AU-coded database has been collected to evaluate the proposed framework, which consists of a "clean" subset containing frontal faces under well controlled circumstances and a challenging subset with large head movements and occlusions. Experiments on this database have demonstrated that the proposed framework yields significant improvement in recognizing speech-related AUs compared to the state-of-the-art visual-based methods especially for those AUs whose visual observations are impaired during speech, and more importantly also outperforms feature-level fusion methods by explicitly modeling and exploiting physiological relationships between AUs and phonemes.

---◆---

## 1 INTRODUCTION

FACIAL behavior is the most powerful and natural means of expressing the affective and emotional states of human being [1]. The Facial Action Coding System (FACS) developed by Ekman and Friesen [2] is a comprehensive and widely used system for facial behavior analysis, where a set of facial *action units* (AUs) are defined. According to the FACS [3], each facial AU is anatomically related to the contraction of a specific set of facial muscles, and combinations of AUs can describe rich and complex facial behaviors. Besides the applications in human behavior analysis, an automatic system for facial AU recognition has emerging applications in advancing human-computer interaction (HCI) such as interactive games, computer-based learning, and entertainment. Extensive research efforts have been focused on recognizing facial AUs from static images or image sequences as discussed in the survey papers [4], [5], [6].

In spite of progress achieved on posed facial display and controlled image acquisition, recognition performance degrades significantly on spontaneous facial displays [7], [8]. Furthermore, it is extremely challenging to recognize AUs that are responsible for producing speech. During speech, these AUs are generally activated at a low intensity with subtle facial appearance/geometrical changes and often introduce ambiguity in detecting other co-occurring AUs [3], e.g., producing non-additive

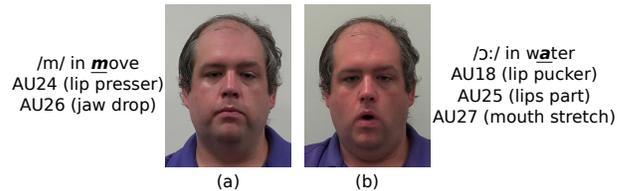

/m/ in **m**ove
AU24 (lip presser)
AU26 (jaw drop)

/ɔ:/ in w**a**ter
AU18 (lip pucker)
AU25 (lips part)
AU27 (mouth stretch)

(a)                    (b)

Fig. 1: Examples of speech-related facial activities, where different AUs are activated non-additively to pronounce speech. (a) The gap between teeth is occluded by the pressed lips in a combination of AU24 and AU26 when sounding /m/ and (b) the space between teeth is partially occluded due to the protruded lips in a combination of AU18, AU25, and AU27 when producing /ɔ:/.

appearance changes. For instance, as illustrated in Fig. 1(a), recognizing AU26 (jaw drop) from a combination of AU24 (lip presser) and AU26 is almost impossible from visual observations when voicing /m/. The reason is that the gap between teeth, which is the major facial appearance clue to recognize AU26 [3], is invisible due to the occlusion by the pressed lips. In another example, when producing /ɔ:/, as shown in Fig. 1(b), AU27 (mouth stretch) would probably be recognized as AU26 because the opening of mouth is much smaller than that when only AU27 is activated due to the activation of AU18 (lip pucker). The failure in recognition of speech-related AUs is because information is extracted from a single source, i.e., the visual channel. As a result, all speech-related AUs are either represented by a uniform code [3], [7], i.e., AD 50, or totally ignored [8], during speech. However, it is critical to identify and differentiate the AUs that are responsible for producing voice from those for expressing emotion and intention, especially during emotional speech.

Instead of solely improving visual observations of AUs, *this*

---

- *Zibo Meng, Shizhong Han, and Yan Tong are with the Department of Computer Science and Engineering, University of South Carolina, Columbia, SC, 29208 USA.*
  *E-mail: mengz, han38, tongy@email.sc.edu*
- *Ping Liu is currently with International Computer Science Institution, University of California, Berkeley, CA.*
  *E-mail: pingliu@icsi.berkeley.edu*

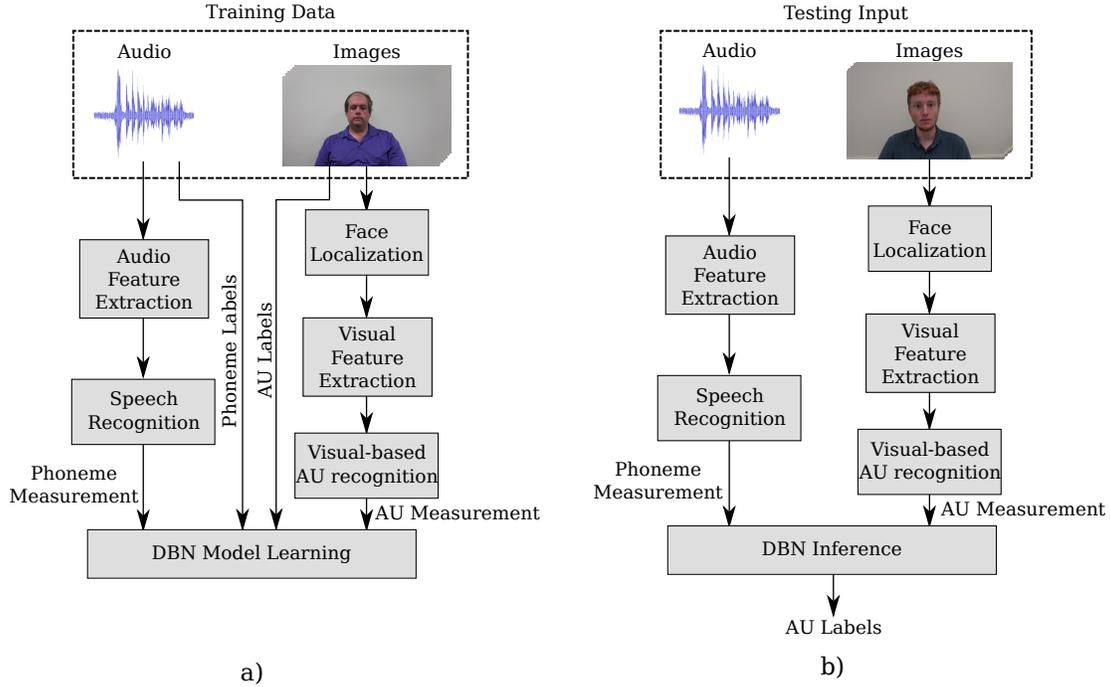

Fig. 2: The flowchart of the proposed audiovisual AU recognition system.

*work aims to explore and exploit the relationships between facial activity and voice to recognize speech-related AUs.* Specifically, facial AUs and voice are highly correlated in two ways. First, voice/speech has strong *physiological relationships* with some lower-face AUs, because jaw and lower-face muscular movements are the major mechanisms for producing different sounds. These relationships are well recognized and have been exploited in natural human communications. For example, without looking at the face, people will know that the other person is opening mouth as hearing "ah". Following the example of recognizing AU26 from a combination of AU24 and AU26 as illustrated in Fig. 1(a), people can easily guess both AU24 and AU26 are activated because of a sound */m/*, although AU26 is invisible from the visual channel. Second, both facial AUs and voice/speech convey human emotions in human communications. Since the second type of relationships is emotion and context dependent, we will focus on studying the physiological relationships between AUs and speech, which are more objective and will generalize better to various contexts.

Since speech can be represented by a sequence of phonemes, each of which is defined as the smallest sound unit in a language, *the relationships between AUs and phonemes will be investigated and explicitly modeled* in this work. Specifically, a phoneme is usually produced by combinations of AUs; and, more importantly, different combinations of AUs are responsible for sounding a phoneme at different phases. For example, */b/* is produced in two consecutive stages, i.e., *stop* and *aspiration*, where AU24 + AU26 and AU25 (lips part) + AU26 are activated, sequentially. Because the physiological relationships between AUs and phonemes are dynamic and stochastic, varying in subjects and languages, we propose to systematically and probabilistically model these relationships by a dynamic Bayesian network (DBN).

Fig. 2 depicts the flowchart of the proposed audiovisual AU recognition system. During training (Fig. 2(a)), a DBN model is learned from the ground truth labels of AUs and phonemes to capture the physiological relationships between AUs and phonemes.

In addition, this DBN model also accounts for measurement uncertainty of AUs and phonemes. For online AU recognition (Fig. 2(b)), AU measurements obtained by visual-based AU recognition and phoneme measurements obtained by speech recognition are employed as evidence for the DBN model. Then, AU recognition is performed by audiovisual information fusion via DBN inference.

In summary, our work has three major contributions.

- A novel audiovisual AU recognition framework is proposed to make the best use of visual and acoustic cues, as humans do naturally.
- Semantic and dynamic physiological relationships between AUs and phonemes are systematically modeled and explicitly exploited by a DBN model to improve AU recognition. To the best of our knowledge, it is the first time to integrate audio information into facial activity analysis by explicitly modeling their relationships.
- An AU-coded audiovisual database has been constructed to evaluate the proposed framework and can be employed as a benchmark database to facilitate audiovisual facial activity analysis.

Experimental results on the new audiovisual database demonstrate that the proposed framework yields significant improvement in recognizing speech-related AUs compared to the state-of-the-art visual-based methods as well as feature-level audiovisual fusion. The performance improvement is more impressive for those AUs, whose visual observations are severely impaired during speech.

## 2 RELATED WORK

### 2.1 Visual based Facial AU Recognition

As discussed in the survey papers [4], [5], [6], most existing approaches for facial AU recognition employed either spatial or temporal features extracted from static images or videos to capture the visual appearance or geometry changes caused by a target AU or AU combinations.

### 2.1.1 Human-designed Facial Features

General purpose human-crafted features are widely employed for facial activity analysis. These features include magnitudes of multi-scale and multi-orientation Gabor wavelets extracted either from the whole face region or at a few fiducial points [9], [10], [11], [12], [13], [14], [15], Haar wavelet features [13] considering the intensity difference of adjacent regions, and Scale Invariant Feature Transform (SIFT) features [16] extracted at a set of keypoints that are invariant to uniform scaling and orientation. Histograms of features extracted from a predefined facial grid have been also employed such as histograms of Local Binary Patterns (LBPs) [17], [18], [19], Histograms of Oriented Gradients (HOG) [20], histograms of Local Phase Quantization (LPQ) features [21], and histograms of Local Gabor Binary Patterns (LGBP) [8], [22]. In addition, spatiotemporal extensions of the aforementioned 2D features, such as LBP-TOP [23], LGBP-TOP [24], [25], LPQ-TOP [21], HOG-TOP [26], and dynamic Haar-like features [27], [28], have been employed to capture the spatiotemporal facial appearance changes caused by AUs.

### 2.1.2 Facial Features Learned from Data

In addition to the human-crafted features, features can also be learned in a data-driven manner by sparse coding or deep learning. As an over-complete representation learned from given input, sparse coding [29] can capture a wide range of variations that are not targeted to a specific application and has achieved promising results in facial expression recognition [30], [31], [32], [33], [34]. By taking advantages of both sparse coding [29] and Non-negative Matrix Factorization (NMF) [35], Nonnegative Sparse Coding (NNSC) [36] has been demonstrated to be effective in facial expression recognition [37], [38], [39], [40]. To become more adaptable to the real world that consists of combination of edges [41], deep learning has been employed for facial expression recognition including deep belief network based approaches [42], [43], [44], [45] and convolutional neural network (CNN) based approaches [46], [47], [48], [49], [50], [51], [52], [53], [54], [55], [56], [57], [58], [58], [87]. Most of these deep-learning based methods took the whole face region as input and learned the high-level representations through a set of processing layers.

All the aforementioned visual-based approaches extracted information solely from the visual channel, and thus are inevitably challenged by imperfect image/video acquisition due to pose variations, occlusions, and more importantly, by the non-additive effects as illustrated in Fig. 1 in recognizing speech-related AUs.

## 2.2 Audio-based Facial AU recognition

Most recently, facial activity recognition from the audio channel has been briefly studied in [59], [60], [61]. Lejan et al. [59] detected three facial activities, i.e. eyebrow movement, smiling, and head shaking, using acoustic information. Assuming that these facial activities are not correlated, different groups of low-level acoustic features are extracted for each facial activity, respectively. Ringeval et al. [60] utilized low-level acoustic feature sets, i.e. ComParE and GeMAPS, for predicting facial AUs for emotion recognition. Our early work [61] employed Mel-Frequency Cepstral Coefficients (MFCC) features extracted from the audio channel for speech-related facial AU recognition. These methods only utilized low-level acoustic features without considering the semantic and dynamic relationships between facial activity and voice. As shown in our previous work [61], AU recognition using low-level acoustic features performed worse than the visual-based approaches for most of the speech-related AUs.

## 2.3 Audiovisual Information Fusion

The proposed framework takes advantage of information fusion of both visual and audio channels, and thus is also related to audiovisual information fusion, which has been successfully demonstrated in automatic speech recognition (ASR) [62], [63] and audiovisual affect/emotion recognition [5]. In the following, we will present a brief review on audiovisual affect/emotion recognition. There are three typical ways to perform audiovisual information fusion.

*Feature-level fusion* directly employs audio and visual features as a joint feature vector for affect/emotion recognition [5], [64]. Recently, deep learning has been employed for learning features from both visual and audio input [58], [65]. In our previous work [61], two feature-level fusion methods were developed for speech-related facial AU recognition. Specifically, one method combined LBP and MFCC features selected from AdaBoost independently; and the other one integrated visual features learned by a CNN with MFCC features. However, these feature-level fusion methods often suffer from differences in time scales, metric levels, and noise levels in the two modalities [5].

*Model-level fusion* [26], [66], [67], [68], [69], [70], [71], [72] exploits correlation between audio and visual channels [5] and is usually performed in a probabilistic manner. For example, coupled [72], tripled [66] or multistream fused HMMs [67], [71] were developed by integrating multiple component HMMs, each of which corresponds to one modality, e.g., audio or visual, respectively. Fragpanagos et al. [68] and Caridakis et al. [69] used an ANN to perform fusion of different modalities. Sebe et al. [70] employed Bayesian network to recognize expressions from audio and facial activities. Chen et al. [26] employed Multiple Kernel Learning (MKL) to find an optimal combination of the features from two modalities.

Most of the existing feature-level fusion or model-level fusion approaches utilize only the low-level features from each modality, e.g. prosody [64], [67], [72], MFCC [61], [64], [72] and formants [64]) for audio channel.

*Decision-level fusion* combines recognition results from two modalities assuming that audio and visual signals are conditionally independent of each other [5], [55], [56], [57], [58], [73], [74], while there are strong semantic and dynamic relationships between audio and visual channels.

In contrast to the major stream of visual-based facial AU recognition, we propose a novel framework for AU recognition by utilizing information from both visual and audio channels. Furthermore, the proposed work differs from the aforementioned audiovisual fusion methods by explicitly capturing and modeling the semantic and dynamic physiological relationships between voice and facial activity. These comprehensive relationships are crucial to describe natural human behaviors, but are neglected in the existing audiovisual fusion approaches.

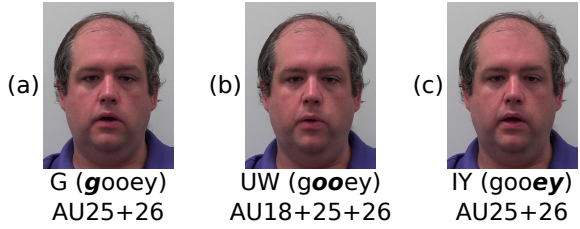

Fig. 3: Examples of the semantic physiological relationships between phonemes and AUs. To pronounce a word *gooey*, different combinations of AUs are activated successively.

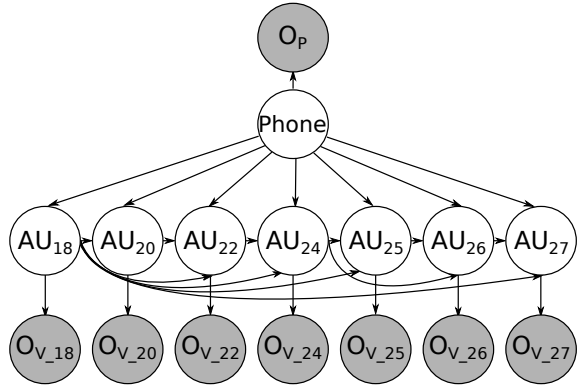

Fig. 4: A BN models semantic physiological relationships between AUs and phonemes as well as the relationships among AUs.

# 3 AUDIOVISUAL FUSION FOR SPEECH-RELATED FACIAL ACTION UNIT RECOGNITION

## 3.1 Modeling Semantic Physiological Relationships between Phonemes and AUs

A phoneme is defined as the smallest sound unit in a language. In this work, a phoneme set defined by the CMU pronouncing dictionary (CMUdict) [75] is employed, which is a machine-friendly pronunciation dictionary developed for speech recognition and describes North American English words using 39 phonemes. According to the CMUdict, the phoneme set and example words, given in the parenthesis, are listed as follows: *AA* (*o*dd), *AE* (*a*t), *AH* (h*u*t), *AO* (*aw*ful), *AW* (c*ow*), *AY* (h*i*de), *B* (*b*e), *CH* (*ch*eese), *D* (*d*ee), *DH* (*th*ee), *EH* (*E*d), *ER* (h*ur*t), *EY* (*a*te), *F* (*f*ee), *G* (*g*reen), *HH* (*h*e), *IH* (*i*t), *IY* (*ea*t), *JH* (*g*ee), *K* (*k*ey), *L* (*l*ee), *M* (*m*e), *N* (*kn*ee), *NG* (pi*ng*), *OW* (*oa*t), *OY* (t*oy*), *P* (*p*ee), *R* (*r*ead), *S* (*s*ea), *SH* (*sh*e), *T* (*t*ea), *TH* (*th*eta), *UH* (h*oo*d), *UW* (t*wo*), *V* (*v*ee), *W* (*w*e), *Y* (*y*ield), *Z* (*z*ee), *ZH* (seizure) [75]. Since each phoneme is anatomically related to a specific set of jaw and lower-face muscular movements, different combinations of AUs are activated to produce different phonemes. Taking the word *gooey* for example, a combination of AU25 and AU26 is first activated to produce the sound *G* (Fig. 3a), and then a combination of AU18, AU25, and AU26 is activated to sound *UW* (Fig. 3b). Finally, AU18 is released and AU25 and AU26 are responsible for sounding *IY* (Fig. 3c).

These semantic physiological relationships are stochastic and vary among subjects. For example, AU20 (lip stretcher) is responsible for producing *AE* in *a*t based on Phonetics [76]; while some subjects do not activate AU20 in practice as found in our audiovisual AU-coded database. In addition, due to the noise in both visual and audio channels, the measurements of AUs and phonemes are not perfect. Instead of employing direct mappings from phonemes to AUs, we propose to use a Bayesian network (BN) to model these semantic relationships probabilistically.

Specifically, each target AU is associated with a node having two discrete states $\{0, 1\}$ representing its *absence* or *presence* status. Phonemes are defined as unique acoustic events and are mutually exclusive at the same time. During speech, a set of phonemes are produced sequentially to speak a meaningful word; and the same phoneme would not repeat itself in two consecutive sound events. Thus, we employ a single node $Phone$ with 29 discrete states representing 28 phonemes, which are involved in producing the words in our audiovisual database, plus one *silence* state denoted as *SIL*. Then, the node $Phone$ can be in one of its 29 states at a certain time to ensure the mutually exclusive relationships among the phonemes.

### 3.1.1 Learning Semantic Physiological Relationships

Given a complete training set including the groundtruth labels of AUs and phonemes, a K2 algorithm [77], implemented in the Bayes Net Toolbox (BNT) [78], is employed to learn the semantic physiological relationships between $Phone$ and the AU nodes. For a node $X_i$, the K2 algorithm finds the set of parents, denoted as $Par(X_i)$ by maximizing the following function [77]:

$$f(X_i, Par(X_i)) = \prod_{j=1}^{M_i} \frac{(K_i - 1)!}{(N_{ij} + K_i - 1)!} \prod_{k=1}^{K_i} N_{ijk}! \quad (1)$$

where $K_i$ is the number of all possible states that $X_i$ may take; $M_i$ is the number of all possible configurations of the parents of $X_i$; $N_{ijk}$ denotes the number of instances where $X_i$ is in its $k^{th}$ state and its parents take the $j^{th}$ configuration; and $N_{ij} = \sum_{k=1}^{K_i} N_{ijk}$.

A BN model learned by the K2 algorithm is shown in Fig. 4, where the nodes represent random variables ($Phone$ and AUs) and the directed links between them represent the conditional dependency. Particularly, the links between $Phone$ and the AU nodes capture their semantic physiological relationships. In addition, since AUs are activated in combinations to produce a meaningful sound, the relationships among AUs are also captured in the BN model by the directed links among them.

## 3.2 Modeling Semantic and Dynamic Relationships using A Dynamic Bayesian Network

By studying Phonetics [76], we know that there are strong physiological relationships between AUs and phonemes. More importantly, these relationships also undergo a temporal evolution. *In particular, there are two kinds of dynamic relationships between AUs and phonemes.*

On the one hand, as the facial muscular movements are activated before a sound is generated [79], the probabilities of the AUs being activated increase and reach an apex as the phoneme is fully made, and then decrease while preparing to produce the next phoneme. On the other hand, different combinations of AUs are responsible for producing a single phoneme at different phases. For example, as illustrated in Fig. 5, the phoneme *B* in *be* has two sequential phases. In the first phase, i.e. the *Stop* phase, the lips are pressed together as activating AU24; the upper and the lower front teeth are usually parted as activating AU26; and "the breath is held and compressed" [76] without emitting sound, i.e., the $Phone$ node is in its silence state *SIL*. As a result, the lip movements (AU24 and AU26) occur earlier than the sound can be heard. In the second phase, i.e. the *Aspiration* phase, the lips part by activating AU25 and the compressed breath is released suddenly as releasing AU24 with an audible explosive sound [76]. Therefore, the physiological relationships between $Phone$ and AUs change over time. In addition, since the duration of the *Stop* phase varies across different subjects and different words, these dynamic relationships are stochastic. Such semantic and dynamic relationships can be well captured by extending the BN (Fig. 4) to a DBN, which not only models the temporal evolution of AUs and phonemes but also models the temporal dependencies among them.

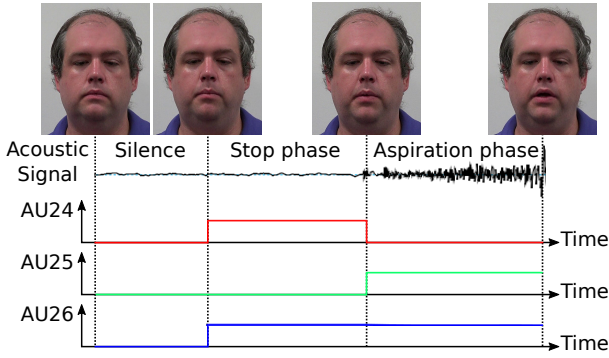

Fig. 5: Illustration of the dynamic relationships between AUs and the phoneme while producing *B* in *be*, where on-axis and off-axis colored lines represent *absence* and *presence* of the corresponding AUs, respectively. Best viewed in color.

### 3.2.1 Learning Dynamic Physiological Relationships

Given a complete set of training data sequences $\mathcal{D} = \{D_1, \ldots, D_S\}$, the dynamic dependencies among AUs and phonemes, i.e., the transition model of a DBN, can be learned by maximizing the following score function

$$Score(B_{tr}) = \log p(\mathcal{D}|B_{tr}) + \log p(B_{tr}) \quad (2)$$

where $B_{tr}$ is a candidate structure of the transition model, and the two terms are the log likelihood and the log prior of $B_{tr}$, respectively. For a large data set, the first term can be approximated using Bayesian information criterion (BIC) [80] as:

$$\log p(\mathcal{D}|B_{tr}) \approx \log p(\mathcal{D}|\hat{\theta}_{B_{tr}}, B_{tr}) - \frac{q}{2}\log S \quad (3)$$

where $\theta_{B_{tr}}$ is a set of model parameters; $\hat{\theta}_{B_{tr}}$ is the maximum likelihood estimation of $\theta_{B_{tr}}$; $q$ is the number of parameters in $B_{tr}$; and $S$ is the number of data samples in $\mathcal{D}$. In Eq. 3, the first term gives the maximum log likelihood of $B_{tr}$ and the second term penalizes the model complexity. In this way, we can learn a DBN model as shown in Fig. 6a.

Specifically, there are two types of dynamic links: self-loops and directed links across two time slices, i.e., from the $(t-1)^{th}$ time slice to the $t^{th}$ time slice. The self-loop at each AU node represents the temporal evolution of each AU; while the self-loop at the $Phone$ node denotes the dynamic dependency between two phonemes. For instance, a consonant is followed by a vowel at the most of time, and hence the probability of a consonant followed by a vowel is much higher than that of a consonant followed by another consonant. The directed links across two time slices characterize the dynamic dependency between two variables, e.g., the dynamic physiological relationships between phonemes and AUs as well as the dynamic relationships among AUs.

**Incorporating domain knowledge in the DBN Structure:** As shown in Fig. 6, the dynamic dependency between AUs in the $(t-1)^{th}$ time slice and $Phone$ in the $t^{th}$ time slice, however, are not learned from data. This is because the penalty, i.e. the second term in Eq. 3, for adding a link from an AU node to $Phone$ is much higher than that from $Phone$ to AU for the 29-state $Phone$ node. Therefore, we refine the learned DBN model by combining the expert knowledge, i.e., the facial muscular movements are activated before sounding a phoneme [79]. Specifically, the dynamic links from AUs to $Phone$ across two successive time slices are manually added as depicted in Fig 6b.

As shown in Fig. 6b, a comprehensive DBN model is constructed. There are two types of nodes in the DBN model: measurement nodes and hidden nodes. The measurement nodes, denoted by the shaded nodes, represent the measurements of AUs denoted by $\mathbf{O}_v$ and the measurement of the phoneme denoted by $\mathbf{O}_p$, whose states can be obtained by visual-based AU recognition and speech recognition, respectively. The hidden nodes are denoted by the unshaded nodes, whose states need to be estimated via probabilistic inference. This DBN model is capable of modeling various interactions in the scenario of audio-visual AU recognition including the semantic and dynamic physiological relationships between AUs and phonemes, semantic and dynamic relationships among AUs, the dynamic relationships between different phonemes, the temporal evolution of AUs, as well as measurement uncertainty.

### 3.3 Learning Model Parameters

Given the model structure as shown in Fig. 6b, the DBN parameters, specified as a set of conditional probabilistic tables (CPTs) associated with each node, can be learned from a set of training data $\mathcal{D} = \{D_1, D_2, \ldots, D_S\}$. The DBN can be considered as an expanded BN consisting of two time slices of static BNs connected by dynamic links. Hence, in addition to learning the CPTs within the same time slice as that does for a static BN, the transition probabilities associated with the dynamic links are also learned. Since the training data is complete in this work, the parameters of the DBN can be estimated using Maximum Likelihood estimation (MLE).

### 3.4 Audiovisual AU Recognition via DBN Inference

Given all available observations from both visual and audio channels until the $t^{th}$ time slice, i.e., $\mathbf{O}_v^{1:t}$ and $\mathbf{O}_p^{1:t}$, AU recognition can be performed through probabilistic inference via the DBN model. Specifically, the posterior probability of the target AUs given all the observations, i.e., $p(\mathbf{AU}^t|\mathbf{O}_v^{1:t}, \mathbf{O}_p^{1:t})$, where $\mathbf{AU}^t$ represents all target AUs at the $t^{th}$ time slice, can be factorized and computed by DBN inference. In this work, a forward-backward inference algorithm implemented in the BNT is employed [78]. Then, the optimal states of the target AUs can be estimated by maximizing the posterior probability:

$$\mathbf{AU}^{t*} = \arg\max_{\mathbf{AU}^t} p(\mathbf{AU}^t|\mathbf{O}_v^{1:t}, \mathbf{O}_p^{1:t}) \quad (4)$$

## 4 Measurement Acquisition

To perform probabilistic inference using the DBN model, the measurements of AUs and the phoneme at each time slice are required. However, signals in different channels are usually sampled at different time scales. For example, the images are sampled at 60 frame per second (fps) and audio tracks are continuous in our audiovisual database. Here, we show how to get AU measurements and phoneme measurements and how to align the measurements from two channels frame by frame.

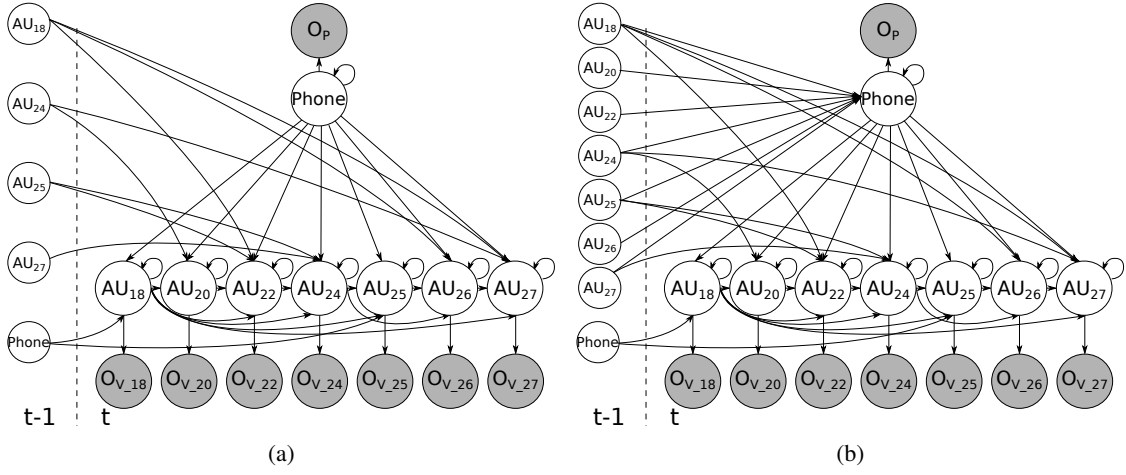

(a)                                          (b)

Fig. 6: A DBN model for audiovisual AU recognition: (a) the DBN structure learned from data, and (b) the DBN structure by integrating expert knowledge into the learned structure. Shaded nodes are the measurement nodes for the corresponding AU nodes and the phoneme node *Phone*, respectively. The links between the unshaded nodes and the shaded nodes characterize the measurement uncertainty.

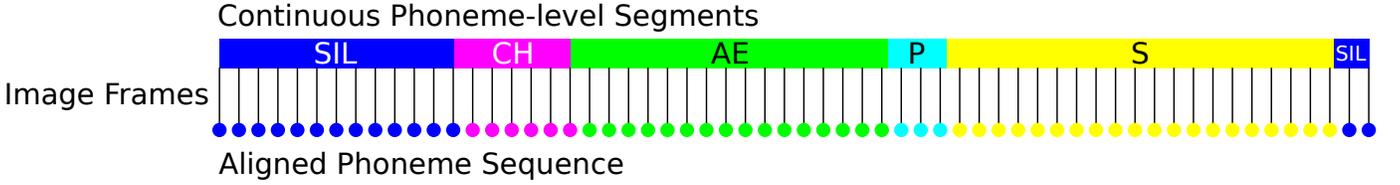

Fig. 7: An illustration of discretizing continuous phoneme segments into frame-by-frame phoneme measurements for the word *chaps*. The top row gives the phoneme-level segments obtained by Kaldi [81]. The last row depicts the aligned sequence of phoneme measurements. The vertical lines in-between represent a sequence of image frames, to which the phonemes will be aligned. Best viewed in color.

### 4.1 Extracting AU Measurements

AU measurements can be obtained using any advanced visual-based facial AU recognition method. In this work, two state-of-the-art visual-based AU recognition methods, i.e., LBP-based method [8] and LPQ-based method [21] are employed to extract AU measurements. Then, audiovisual information fusion is performed based on the two types of AU measurements, respectively.

#### 4.1.1 Extracting LBP features

For preprocessing purposes, the face regions across different facial images are aligned to remove scale and positional variance [82] and then cropped to $96 \times 64$, which are further divided into $7 \times 7$ grid. From each grid, LBP histograms with 59 bins are extracted and then, concatenated into a single vector, which is denoted as LBP feature.

#### 4.1.2 Extracting LPQ features

The facial images are preprocessed following the same procedure as described in Section 4.1.1 and divided into $7 \times 7$ grid. From each grid, LPQ histograms with 256 bins are extracted and then, concatenated into a single vector, which is denoted as LPQ feature.

For each target AU, an AdaBoost classifier is employed to select the most discriminative features from either LBP or LPQ feature pool and construct a strong classifier to perform facial AU recognition for each target AU. The binary classification results obtained from AdaBoost classifiers will be fed into the DBN model as the AU measurements. Hence, we have two types of AU measurements based on the two kinds of visual features, respectively. Note that, any visual-based AU recognition method can be adopted to extract measurements of AUs.

### 4.2 Extracting Phoneme Measurements

In this work, a state-of-the-art speech recognition method, i.e., Kaldi toolkit [81], is employed to obtain the phoneme measurements. Specifically, 13-dimensional MFCC [83] features are extracted and employed in Kaldi to get word-level speech recognition results, which are further divided into phoneme-level segments as shown in Fig. 7. In order to obtain a phoneme measurement for each time slice, which should be also synchronized with the AU measurements, the continuous phoneme segments are discretized according to the sampling rate of the image frames, i.e., 60 fps in our experiment. As illustrated in Fig. 7, the first row shows the continuous phoneme-level segments for the word *chaps*; the second row shows a sequence of image frames to be aligned to; and the last row shows the frame-by-frame phoneme measurements, which are synchronized with the image frames and will be fed into the DBN model as the measurements for *Phone* for audiovisual AU recognition.

## 5 EXPERIMENTAL RESULTS

### 5.1 Audiovisual AU-coded Dataset

As far as we know, the current publicly available AU-coded databases only provide information in visual channel. Further-more, all the speech-related AUs have been either annotated by a uniform label, i.e., AD 50 [7] or not labeled [8], during speech. In order to learn the semantic and dynamic physiological relationships between AUs and phonemes, as well as to demonstrate the proposed audiovisual AU recognition framework, we have constructed a pilot AU-coded audiovisual database consisting of two subsets, i.e. a *clean* subset and a *challenging* subset. Fig. 8 illustrates example images of the speech-related AUs in the audiovisual database.

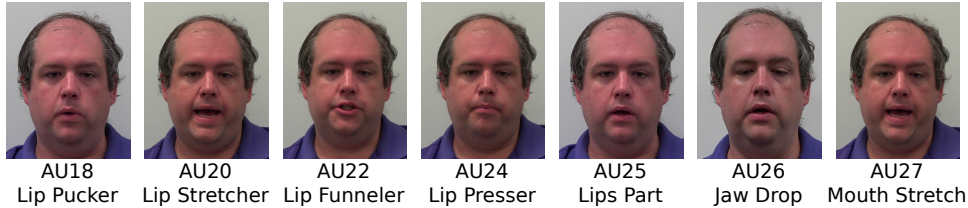

| AU18 | AU20 | AU22 | AU24 | AU25 | AU26 | AU27 |
|------|------|------|------|------|------|------|
| Lip Pucker | Lip Stretcher | Lip Funneler | Lip Presser | Lips Part | Jaw Drop | Mouth Stretch |

Fig. 8: A list of speech-related AUs and their interpretations in the audiovisual database.

In the audiovisual database, there are a total of 13 subjects of different races, ages, and genders, where 2 subjects appear in both the *clean* and *challenging* subsets. The database consists of 12 words [1], which contain 28 phonemes and the most representative relationships between AUs and phonemes. All the videos in this database were recorded at 59.94 frames per second at a spatial resolution of $1920 \times 1080$ with a bit-depth of 8 bits; and the audio signals were recorded at 48kHz with 16 bits. The statistics, i.e., the numbers of occurrences, of the speech-related AUs in the *clean* and *challenging* subsets are reported in Table. 1.

TABLE 1: Statistics of the speech-related AUs in the audiovisual database.

| Subsets | AU18 | AU20 | AU22 | AU24 | AU25 | AU26 | AU27 | Total Frames |
|---------|------|------|------|------|------|------|------|--------------|
| Clean | 7,014 | 1,375 | 4,275 | 2,105 | 25,092 | 18,280 | 4,444 | 34,622 |
| Challenging | 4,118 | 1,230 | 3,396 | 1,373 | 17,554 | 11,830 | 3,242 | 23,274 |

***Clean* subset:** Videos were collected from 9 subjects. Each subject was asked to speak the 12 words individually, each of which will be repeated 5 times. In addition, all subjects were required to keep a neutral face during data collection to ensure all the facial activities are only caused by speech.

***Challenging* subset:** Videos were collected from 6 subjects speaking the same words 5 times as those in the *clean* subset. As illustrated in Fig. 9, the subjects were free to display any expressions on their face during speech and were not necessary to show neutral face before and after speaking the word. In addition, instead of being recorded from the frontal view, videos were collected mostly from the sideviews with free head movements and occlusions by glasses, caps, and facial hair, introducing challenges to AU recognition from the visual channel. In addition, the location of microphone varies for different subjects, introducing challenges to the audio channel.

Groundtruth phoneme segments and AU labels were annotated in the database. Specifically, the utterances were transcribed using the Penn Phonetics Lab Forced Aligner (p2fa) [84], which takes

<hr/>

1. The 12 words including "beige", "chaps", "cowboy", "Eurasian", "gooey", "hue", "joined", "more", "patch", "queen", "she", and "waters" were selected from English phonetic pangrams (http://www.liquisearch.com/list_of_pangrams/english_phonetic_pangrams) that consists of all the phonemes at least once in 53 words.

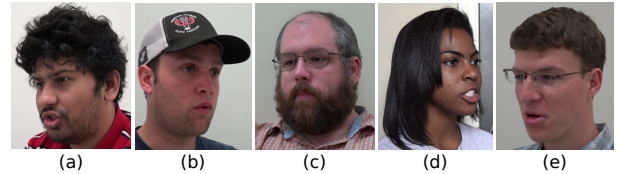

Fig. 9: Example images in the *challenging* subset collected from different illuminations, varying view angles, and with occlusions by glasses, caps, or facial hairs.

an audio file along with its corresponding transcript as input and produces a Praat [85] TextGrid file containing the phoneme segments. 7 speech-related AUs, i.e. AU18, AU20, AU22, AU24, AU25, AU26, and AU27, as shown in Fig. 8, were frame-by-frame labeled manually by two certified FACS coders. Roughly 10% of the data was labeled by both coders independently to estimate inter-coder reliability measured by Matthews Correlation Coefficient (MCC) [86]. As illustrated in Table 2, the MCC for each AU ranges from 0.69 for AU27 to 0.98 for AU25 and has an average of 0.88 on the *clean* subset, and ranges from 0.80 for AU26 to 0.96 for AU25 on the *challenging* subset, which indicates strong to very strong inter-coder reliability of AU annotation.

TABLE 2: Inter-coder reliability measured by MCC for the 7 speech-related facial AUs on the audiovisual database.

| Subsets | AU18 | AU20 | AU22 | AU24 | AU25 | AU26 | AU27 | Total Frames |
|---------|------|------|------|------|------|------|------|--------------|
| Clean | 0.945 | 0.930 | 0.864 | 0.944 | 0.985 | 0.791 | 0.695 | 0.879 |
| Challenging | 0.941 | 0.917 | 0.930 | 0.824 | 0.963 | 0.799 | 0.842 | 0.888 |

### 5.2 Methods in Comparison

Experiments have been conducted on the new audiovisual AU-coded dataset to demonstrate the effectiveness of the proposed framework, as shown in Fig. 6b, in improving the recognition performance for speech-related AUs. In this work, we built the proposed approach upon two state-of-the-art visual-based methods, i.e. LBP-based and LPQ-based methods, denoted as *DBN-LBP* and *DBN-LPQ*, respectively. Each method is first compared with the state-of-the-art visual-based methods utilizing the same features. Furthermore, they are compared with three baseline audiovisual fusion methods to demonstrate the effectiveness of explicitly modeling and employing the semantic and dynamic physiological relationships between AUs and phonemes in audiovisual fusion. The LBP-based baseline methods are described as follows.

**Ada-LBP:** A state-of-the-art LBP-based AU recognition approach [8], denoted as *Ada-LBP*, is employed as a static visual-based baseline approach, as described in Section 4.

**DBN-V-LBP:** A state-of-the-art DBN-based model [14], denoted as *DBN-V-LBP*, is employed to model the relationships among AUs and used as a dynamic visual-based baseline. The structure of *DBN-V-LBP* is obtained by eliminating the "**Phone**" node and its measurement node from the *DBN-LBP* structure depicted by Fig. 6a.

The comparison between the visual-based baselines, i.e., *Ada-LBP* and *DBN-V-LBP*, and all audiovisual fusion methods aims to demonstrate the effectiveness of incorporating information from audio channel in improving AU recognition.

**Ada-Fusion-LBP:** The first baseline fusion method, denoted as *Ada-Fusion*, was developed and described in our early work [61], which employs a *feature-level fusion scheme* that extracts features from both visual and audio channels, i.e. histograms of LBP features and MFCC features. Specifically, given an input wave file, MFCCs are extracted using window size $l = 16.67ms$ with a frame-shift $s = 16.67ms$ by Kaldi toolkit [81]. To include more temporal information, 7 frames, i.e. 3 frames before and after the current frame along with the current one, are concatenated as the final MFCC feature for each frame. The extracted LBP and MFCC features are integrated into a single feature vector and employed as the input for AdaBoost to make predictions for the target AU. The comparison between the *Ada-Fusion* and the other probabilistic-model-based fusion methods intends to show the effectiveness of explicitly modeling the physiological relationships between AUs and phonemes.

**BN-LBP** The second baseline fusion method, denoted as *BN-LBP*, employs a static BN model with a structure illustrated in Fig. 4 plus measurement nodes for all AUs and the $Phone$ node. The *BN-LBP* only considers the semantic relationships between AUs and phonemes as well as the semantic relationships among AUs, while the dynamics of AUs and phonemes are ignored. The comparison between *BN-LBP* and the DBN-based models is used to demonstrate the importance of modeling the dynamic physiological relationships between AUs and phonemes as well as their temporal evolution.

**DBN-Learned-LBP** The last baseline fusion method, denoted as *DBN-learned-LBP*, employs the learned DBN structure model as shown in Fig. 6a and does not model the dynamic dependencies between AUs in the $(t-1)^{th}$ time slice and phonemes in the $t^{th}$ time slice. The comparison between *DBN-learned-LBP* and the proposed *DBN-LBP* is employed to demonstrate the effectiveness of integrating expert knowledge into model learning.

The LPQ-based baseline methods are defined in the same way, whose model structures are the same as those of the LBP-based equivalents. For all methods evaluated, a leave-one-subject-out training/testing strategy is employed, where the data from one subject is used for testing and the remaining data is used for training.

### 5.3 Experimental Results and Data Analysis on the Clean Subset

We first evaluate the proposed *DBN-LBP* and *DBN-LPQ* on the *clean* subset.

#### 5.3.1 Experimental Results of LBP-based methods

Quantitative experimental results of LBP-based methods are reported in Fig. 10 in terms of F1 score, false positive rate (FPR), and true positive rate (TPR) for the 7 speech-related AUs. As shown in Fig. 10, all the audiovisual fusion methods outperform the static visual-based method, i.e. *Ada-LBP*. Specifically, the overall recognition performance is improved from **0.416** using the *Ada-LBP* to **0.463** (*Ada-Fusion-LBP*), **0.658** (*BN-LBP*), **0.666** (*DBN-Learned*), and **0.696** by the proposed *DBN-LBP*, in terms of the average F1 score, which demonstrates that information from the audio channel indeed helps the recognition of speech-related AUs.

All the fusion methods, except the feature level fusion method, i.e. *Ada-Fusion-LBP*, perform better than the dynamic visual-based method, i.e. *DBN-V-LBP*. The performance of *Ada-Fusion-LBP* is inferior to that of *DBN-V-LBP* because subjects in the *clean* subset were asked to produce the words and display the lip movements clearly, and thus the relationships between AUs, explicitly modeled by *DBN-V-LBP*, are strong. Furthermore, as shown in Fig. 10, the proposed *DBN-LBP* framework outperforms all methods compared in terms of the average F1 score (**0.696**), the average FPR (**0.071**), and the average TPR (**0.732**). In the following, we will compare the proposed *DBN-LBP* with each baseline method side by side.

**Comparison of *Ada-LBP*, *DBN-V-LBP*, and *DBN-LBP*** The proposed *DBN-LBP* model significantly outperforms the state-of-the-art visual-based method *Ada-LBP* and *DBN-V-LBP* on all target AUs. Notably, *DBN-LBP* drastically improves the recognition performance of AU26 (jaw drop) and AU27 (mouth stretch). For example, the F1 score of AU27 increases from **0.273** (*Ada-LBP*) and **0.328** (*DBN-V-LBP*) to **0.720** (*DBN-LBP*). The performance improvement is primarily because of the integration of audio information in recognition of these AUs. As shown in Fig. 1, the visual cues for recognizing AU26 and AU27 are severely impaired by occlusions introduced by the presence of other AUs during speech. However, the information from the audio channel is not affected and thus, more reliable.

**Comparison between *Ada-Fusion-LBP* and *DBN-LBP*** As shown in Fig. 10, *DBN-LBP* outperforms *Ada-Fusion-LBP*, a feature-level fusion method, for all target AUs. Specifically, the average F1 score is improved from **0.463** (*Ada-Fusion*) to **0.696** (*DBN-LBP*). The performance improvement mainly comes from two aspects. First, the proposed *DBN-LBP* benefits from the remarkable achievements in speech recognition: the speech recognition performance of the Kaldi Toolkit [81] in terms of the word-level error rate is 1.3% on the *clean* subset in our experiments. Hence, it makes more sense to employ accurate phoneme measurements than to use low-level acoustic features directly. Second, it is more effective to explicitly model and exploit the semantic and dynamic physiological relationships between AUs and phonemes than to employ the audiovisual features directly. For example, the TPR of AU26 increases from **0.635** (*Ada-Fusion-LBP*) to **0.730** (*DBN-LBP*) with a drastic decrease in the FPR from **0.338** (*Ada-Fusion-LBP*) to **0.136** (*DBN-LBP*) by employing the physiological relationships between the phonemes and AUs as shown in Fig. 1.

**Comparison between *BN-LBP* and *DBN-LBP*** Both AUs and phonemes are dynamic events and their dynamics are crucial in natural communications. By modeling both the semantic and dynamic relationships between AUs and phonemes, the recognition performance of using *DBN-LBP* is better than that of using *BN-LBP* for all target AUs in terms of all metrics. For example, there are strong dynamic relationships between phonemes and AU24 (lip presser), e.g., AU24 is activated in the *Stop* phase of *B* in *b*e before the sound is emitted, as depicted in Fig. 5. As shown in Fig. 10, the recognition performance of AU24 gains a significant improvement using *DBN-LBP*: the F1 score increases from **0.394** (*BN-LBP*) to **0.560** (*DBN-LBP*).

**Comparison between *DBN-Learned-LBP* and *DBN-LBP***

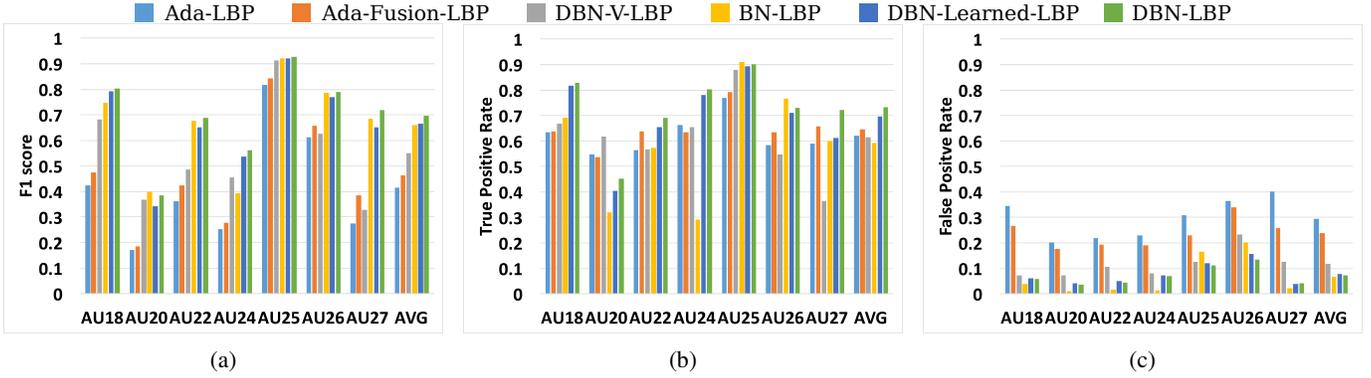

Fig. 10: Performance comparison of AU recognition on the clean subset in terms of (a) F1 score, (b) true positive rate, and (c) false positive rate.

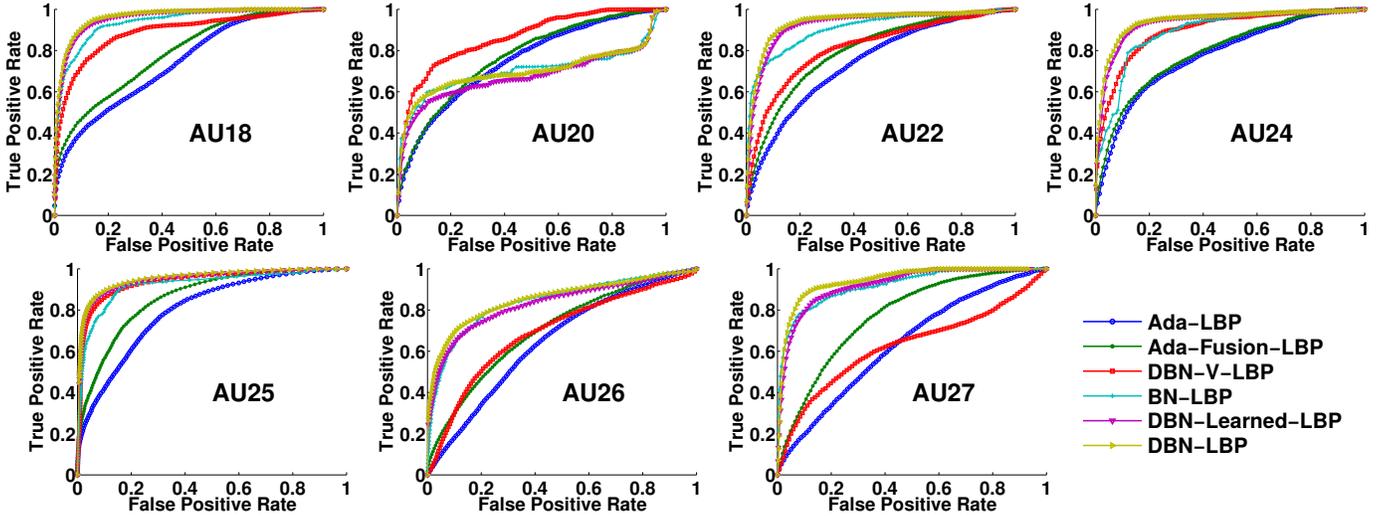

Fig. 11: ROC curves for 7 speech-related AUs on the clean subset using LBP features. Best viewed in color.

Since AUs are the major mechanism to produce the voice, they are activated before the sound is produced [79]. As shown in Fig. 10, the *DBN-LBP* outperforms the *DBN-Learned-LBP* for all target AUs in terms of F1 score, which demonstrates the effectiveness of integrating this expert knowledge into the learned DBN model.

In addition to the three metrics, an ROC analysis is conducted for each AU to further demonstrate the performance of the proposed framework. As show in Fig. 11, each ROC curve is obtained by plotting the TPR against FPR at different thresholds over the predicted scores. The performance of the proposed *DBN-LBP* model is better or at least comparable with that of the baseline methods on all the target AUs except for AU20. As shown in Fig. 11, the performance of the fusion-based methods is inferior to that of the dynamic visual-based method, i.e., *DBN-V-LBP*, because the relationships between AU20 and the phonemes are weak due to large variations among subjects. For example, although AU20 is responsible to producing the phoneme AE in *chaps* according to Phonetics [76], some subjects did not activate AU20 during speech.

Furthermore, Fig. 12 gives an example of the system outputs, i.e., the estimated probabilities of AUs, by *DBN-LBP* and *DBN-Learned-LBP*, respectively. As shown in Fig. 12, when sounding a word *chaps*, the probabilities of AUs increase when they are preparing to sound a phoneme and decrease rapidly after the sound is emitted. As the facial movements are activated before the sound is generated, the probability of AU22 increases and reaches above 0.5, i.e. the activation threshold, before the phoneme *CH* is made. The solid vertical yellow line represents the onset time point of AU22 labeled manually, while the dashed line shows the estimated activation time of AU22. The closer those two lines are, the better the prediction is. By integrating the expert knowledge, *DBN-LBP* can better predict the activation of an AU given the measurements. In another example, AU27 was not detected by *DBN-Learned-LBP* shown in the bottom plot, because the visual-based classifier fails to detect AU27. However, the dynamic relationships between AUs and phonemes utilized by *DBN-LBP* can help to predict AU27 despite the poor visual measurements.

### 5.3.2 Experimental Results of LPQ-based methods

The quantitative experimental results using LPQ features are reported in Table 3 in terms of the average F1 score. Not surprisingly, the proposed *DBN-LPQ* outperforms all the compared methods in terms of F1 score (**0.679**).

*The performance improvement on both the LBP-based method and the LPQ-based method demonstrates that the proposed method can be built upon any advanced visual-based AU recognition method, and consistently improve the performance for speech-related AUs recognition.*

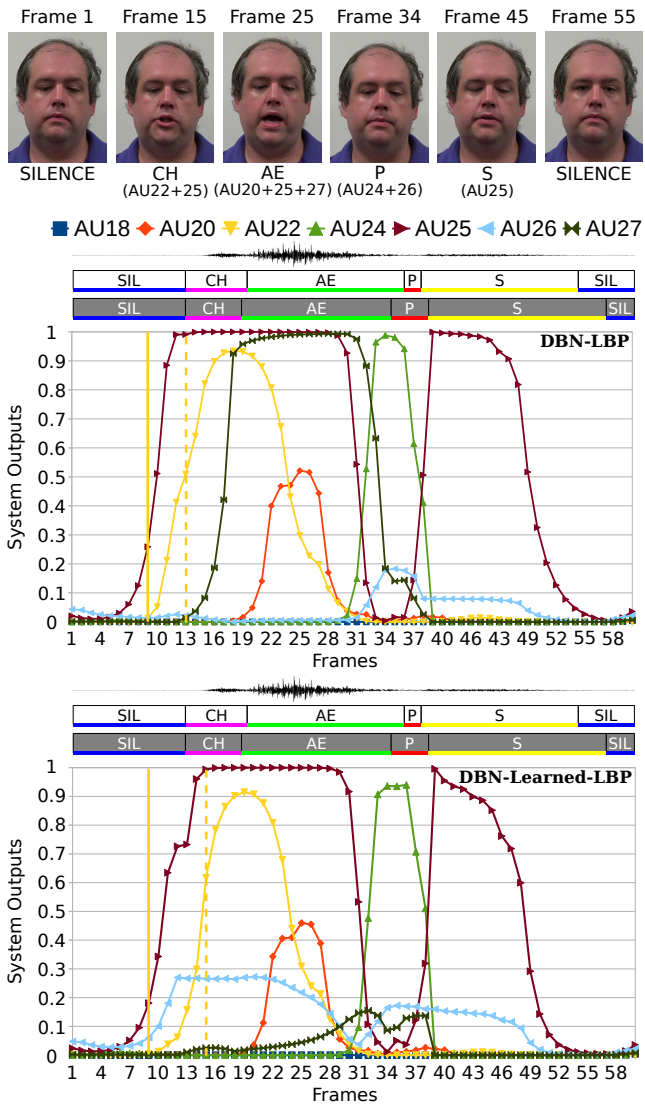

Fig. 12: An example of the system outputs by DBN inference using *DBN-LBP* and *DBN-Learned-LBP*, respectively. A word *chaps* is produced and AU20, AU22, AU24, AU25, and AU27 have been activated. The top row shows key frames from the image sequence, as well as the AU combinations for producing the corresponding phonemes. The two bottom figures depict the probabilities of AUs estimated by *DBN-LBP* and *DBN-Learned-LBP*, respectively. The unshaded phoneme sequence is the ground truth and the shaded one represents the evidence utilized by DBN models. The dashed and solid vertical lines denote the ground truth and the predicted time point, where AU22 is activated, respectively. The dash vertical line is closer to the solid vertical line in *DBN-LBP*. Best viewed in color.

TABLE 3: Performance comparison on the two subsets in terms of the average F1 score.

| Subsets | Ada-LBP | Ada-Fusion-LBP | DBN-V-LBP | BN-LBP | DBN-Learned-LBP | DBN-LBP |
|---|---|---|---|---|---|---|
| **Clean** | 0.416 | 0.463 | 0.551 | 0.658 | 0.666 | **0.696** |
| **Challenging** | 0.372 | 0.448 | 0.368 | 0.608 | 0.548 | **0.622** |
| Subsets | Ada-LPQ | Ada-Fusion-LPQ | DBN-V-LPQ | BN-LPQ | DBN-Learned-LPQ | DBN-LPQ |
| **Clean** | 0.448 | 0.482 | 0.536 | 0.651 | 0.651 | **0.679** |
| **Challenging** | 0.362 | 0.430 | 0.370 | 0.619 | 0.579 | **0.651** |

## 5.4 Experimental Results and Data Analysis on the Challenging Subset

Experiments were conducted on the *challenging* subset to further demonstrate the advantage of incorporating audio information for speech-related AU recognition under real world conditions, where facial activities are accompanied by free head movements and often with occlusions caused by facial hairs, caps, or glasses.

The proposed *DBN-LBP* and *DBN-LPQ* models and the baseline methods were trained and tested on the *challenging* subset using a leave-one-subject-out strategy. Since there are only 6 subjects in the *challenging* subset, we employed the data in the *clean* subset as additional training data, except those of the two subjects who also appear in the *challenging* subset to ensure a subject-independent context. In particular, the data of 5 subjects from the *challenging* subset along with the data of 7 subjects from the *clean* subset is used as the training data, and the remaining one subject from the *challenging* subset is employed as the testing data. The structures of the *DBN-Learned* and *DBN* trained on the *challenging* subset are shown in Fig. 13a and Fig. 13b, respectively.

### 5.4.1 Experimental Results and Discussion

Quantitative experimental results on the *challenging* subset are reported in Fig. 14 for LBP-based methods, in terms of F1 score, TPR, and FPR for the 7 speech-related AUs. From Fig. 14, we can find that all the audiovisual fusion methods outperform the methods employing only visual information (*Ada-LBP* and *DBN-V-LBP*). Specifically, as the head movements and occlusions are introduced in the *challenging* subset, the visual observations of AUs become unreliable, which is reflected by the drastic drop in performance of the visual-based methods, i.e. **0.372** (*Ada-LBP*) and **0.368** (*DBN-V-LBP*) in terms of the average F1 score. In contrast, the information extracted from the audio channel is less affected. Thus, the performance is improved from **0.372** (*Ada-LBP*) to **0.448** (*Ada-Fusion-LBP*), **0.608** (*BN-LBP*), **0.548** (*DBN-Learned-LBP*), and **0.622** by the proposed *DBN-LBP* in terms of average F1 score. Similar to that on the *clean* subset, Fig. 15 gives an example of the system outputs, i.e., the estimated probabilities of AUs, by the proposed *DBN-LBP*.

Experimental results on the two subsets for both LBP-based and LPQ-based methods are reported in Table 3 in terms of the average F1 score. Notably, both *DBN-V-LBP* and *DBN-V-LPQ* have performance comparable to *Ada-LBP* and *Ada-LPQ*, respectively, on the *challenging* subset, since the visual observations become unreliable under spontaneous settings. Moreover, the dynamic dependencies between AUs and phonemes become more important under spontaneous conditions. For example, by comparing Fig. 6a and Fig. 13a, more temporal links between phonemes and AUs are learned under real-world conditions in Fig. 13a. In addition, by incorporating the expert knowledge, the proposed *DBN-LBP* and *DBN-LPQ* improve the performance by **0.074** and **0.072** compared with the *DBN-Learned-LBP* and *DBN-Learned-LPQ*, respectively, in terms of F1 score.

## 5.5 Comparison with More State-of-the-art Visual-based Methods

To further demonstrate the effectiveness of the proposed framework, two more state-of-the-art visual-based methods are implemented and evaluated on the AU-coded audiovisual database using a leave-one-subject-out cross-validation strategy. One of the methods is based on a kind of human-crafted feature, i.e. LGBP features [8], [22], denoted as *LGBP*. Specifically, 400 LGBP features are selected by AdaBoost and employed to train an SVM classifier for each target AU. The other method is based on a deep learning model, i.e. incremental boosting convolutional neural network (IB-CNN) [87]. Since only the lower-part of the face is responsible for producing the speech-related AUs, a two-stream IB-CNN is developed in this work to learn both appearance and shape information of lip regions, denoted as *IB-CNN-LIP*. Particularly, the lip region along with the landmarks on the lip are employed to train the IB-CNN-LIP model. Experimental results on both subsets of all methods in comparison can be found in Table 4 in terms of average F1 score. As illustrated in Table 4, the proposed *DBN-LBP* and *DBN-LPQ* outperform *LGBP* and *IB-CNN-LIP* by a large margin in terms of the average F1 score. The improvement mainly comes from utilizing both acoustic and visual information by explicitly modeling the semantic and dynamic relationships between phonemes and AUs in *DBN-LBP* and *DBN-LPQ*. In contrast, *LGBP* and *IB-CNN-LIP* make predictions based on only visual clues, where the appearance changes for AUs are subtle, and sometimes "invisible" in the visual channel.

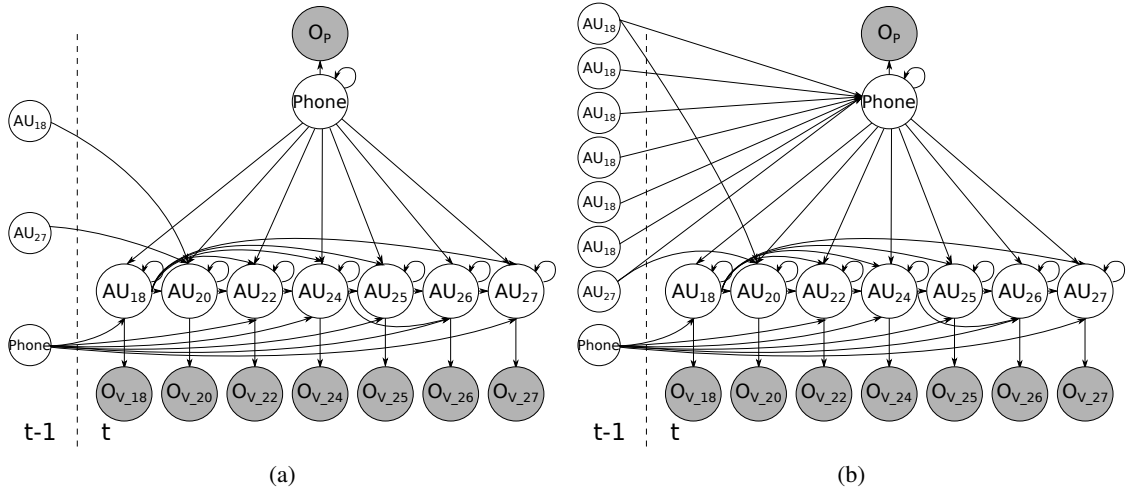

Fig. 13: A DBN model for audiovisual AU recognition: (a) the DBN structure learned from the challenging data, and (b) the DBN structure by integrating expert knowledge into the learned structure.

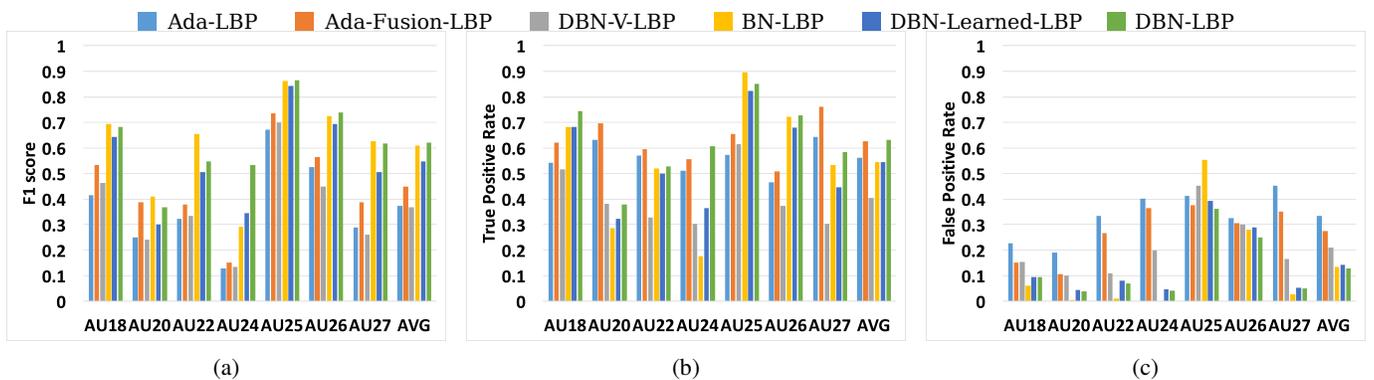

Fig. 14: Performance comparison of AU recognition on the challenging subset in terms of (a) F1 score, (b) true positive rate, and (c) false positive rate.

TABLE 4: Performance comparison on the two subsets in terms of the average F1 score with two state-of-the-art methods.

| Subsets | *DBN-LBP* | *DBN-LPQ* | *LGBP* [8] | *IB-CNN-LIP* [87] |
|---|---|---|---|---|
| **Clean** | 0.696 | 0.679 | 0.391 | 0.465 |
| **Challenging** | 0.622 | 0.651 | 0.336 | 0.382 |

## 6 CONCLUSION AND FUTURE WORK

Facial activity is not the only channel for human communication, where voice also plays an important role. This paper presents a novel audiovisual fusion framework for recognizing speech-related AUs by exploiting information from both visual and audio channels. Specifically, a DBN model is employed to capture the comprehensive relationships for audiovisual AU recognition including the semantic and dynamic relationships among AUs, the temporal development of AUs, the dynamic dependencies among phonemes, and more importantly, the semantic and dynamic physiological relationships between phonemes and AUs. Experimental results on a new audiovisual AU-coded dataset have demonstrated that the proposed DBN framework significantly outperforms the state-of-the-art visual-based methods by incorporating audio information in facial activity analysis. Furthermore, the DBN model also beat the feature-level fusion by comprehensively modeling and exploiting the relationships in a context of audiovisual fusion.

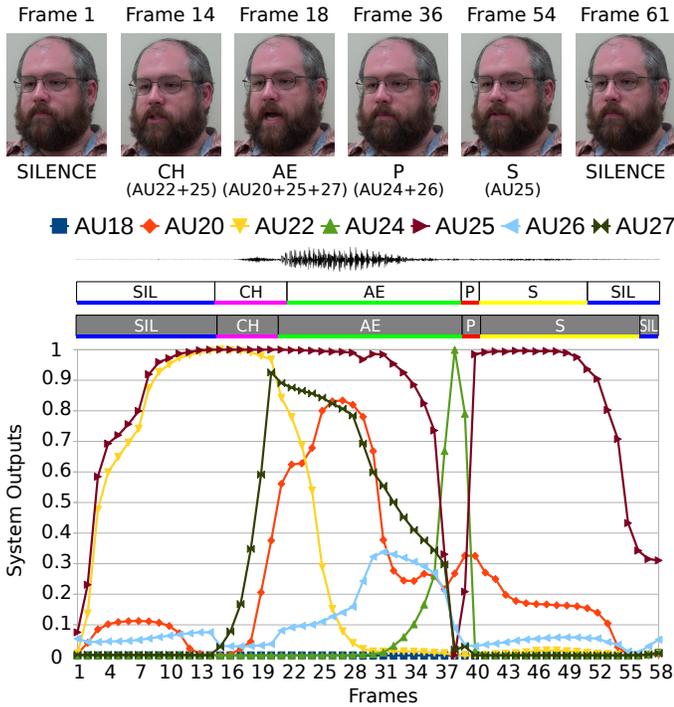

Fig. 15: An example of the system outputs of *DBN-LPB* on the challenging subset, where a word *chaps* is produced and AU20, AU22, AU24, AU25, and AU27 are activated. The top row shows key frames from the image sequence as well as the corresponding AU combinations. The bottom figure depicts the probabilities of AUs changing over time. The unshaded phoneme sequence is the ground truth phoneme labels and the shaded one is the evidence of the *DBN-LPB*. Best viewed in color.

In the future, we plan to enrich the pilot audiovisual database with more challenging data including emotional speech, and then extend the current framework to modeling more complicated relationships in natural human communications.

## ACKNOWLEDGMENT

This work is supported by National Science Foundation under CAREER Award IIS-1149787.

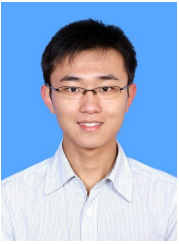

**Zibo Meng** received a Master degree from Zhejiang University, China, in 2013. He is currently pursuing his Ph.D. degree at University of South Carolina, Columbia, South Carolina. His areas of research include computer vision, pattern recognition, and information fusion. He is a student member of the IEEE.

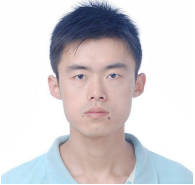

**Shizhong Han** is pursuing his Ph.D. degree at University of South Carolina, Columbia, South Carolina. He received a Master degree from Wuhan University, China, in 2010. His research interests include computer vision, pattern recognition, machine learning, and data mining. He is a student member of the IEEE.

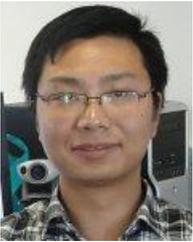

**Ping Liu** received his Ph.D. degree under the supervision from Prof. Yan Tong, Department of Computer Science and Engineering, University of South Carolina. Before joining University of South Carolina, he was a research assistant in the Intelligent Systems and Biomedical Robotics Group (ISBRG), University of Portsmouth, UK. He got his Master Degree from Institute for Pattern Recognition and Artificial Intelligence (IPRAI), Huazhong University of Science and Technology (HUST), WuHan, China; Bachelor Degree (EE) from Wuhan University of Technology (WUT), WuHan, China. His research interests include computer vision and machine learning. He is a member of the IEEE.

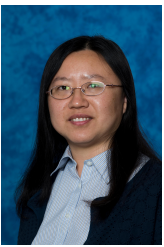

**Yan Tong** received the Ph.D. degree in electrical engineering from Rensselaer Polytechnic Institute, Troy, New York, in 2007. She is currently an associate professor in the Department of Computer Science and Engineering, University of South Carolina, Columbia, SC, USA. From 2008 to 2010, she was a research scientist in the Visualization and Computer Vision Lab of GE Global Research, Niskayuna, NY. Her research interests include computer vision, machine learning, and human computer interaction. She has served as a conference organizer and a program committee member for a number of premier international conferences. She is a member of the IEEE.